# 'Just Enough' Ontology Engineering

Paola Di Maio
University of Strathclyde/ISTCS.org
75 Montrose Street
Glasgow UK

paola.dimaio[@gmail.com

## ABSTRACT
This paper introduces 'just enough' principles and 'systems engineering' approach to the practice of ontology development to provide a minimal yet complete, lightweight, agile and integrated development process, supportive of stakeholder management and implementation independence.

## 1. INTRODUCTION
The majority of Information centric systems today are designed to leverage knowledge expressed via natural language, where symbols and meanings (semiotics and semantics) need to be captured and represented adequately for these systems to function. Ontologies are conceptual and semantic representations widely used, in different forms, to capture and express such models. Ontology Engineering Methodologies (also called ontology Development by some) have proliferated in recent years, however how to go about developing an ontology for a general project or organisation can be very resource intensive, require skills and expertise that are 'scarce', and can be a minefield of uncertainties. A plethora of ontology engineering (OE) methods, artifacts, tools and techniques has surfaced in recent years, often resulting from academic research dissemination, but OE has not become any easier. The choice of an appropriate ontology development methodology for any project may require a systematic evaluation of existing approaches, and this can become extremely resource intensive and time consuming.A default option is to follow no methodology at all (the 'who needs a methodology?' attitude) often preferred by developers who go straight into coding, and may confuse 'on the fly' schema creation with fully fledged ontology development. JEOE incorporates principles of *'just enough*' approaches into OE, with the intent to capture and synthesize essential aspects of OE independently from the methodology of choice. *Just enough ontology engineering* does not intend to be a methodology, nor to substitute for methodologies, rather as an approach to help specialists from various disciplines (and not necessarily experts in model building) who may be called to contribute to conceptual modelling efforts, and to equip them with minimal all around competences and understanding of ontology development. 'High order' OE skills are scarce, but also not particularly useful unless coupled with practical systems development, management, and a good measure of common sense, which unfortunately, is not always the most salient characteristic of pure ontologists and theoreticians.

### 1.1 JEOE in a nutshell
Even the smallest ontology, when well formed and properly 'grounded', can be reused and incorporated into (or at least referenced by) larger ontologies, thus effectively contributing to address one of the contemporary challenges which is facilitating the access, validation, maintenance and reuse of existing Knowledge. JEOE consists mainly of a lean and compact set of steps, an agile process that can be adopted to navigate iteratively across the many interdependent OE activities, essentially providing what anyone with limited time and without formal training in logic, mathematics or philosophy may need to know to contribute intelligently to discourse. JEOE can be thought of as 'ontology development for the rest of us' in a nutshell. It contributes a pragmatic, systemic approach to ontology building which is methodology independent and provides a framework to help incorporate elements of systems development principles into the often challenging practice of ontology development. It also serves as didactic/learning instrument for generalists, and as a tool to guide development across team members with different cognitive and disciplinary backgrounds. Another non trivial problem that JEOE aims to tackle is the lack of publicly available usable documentation resources for existing OE methodologies. Many OE papers, especially when authored by academics, may not be designed for practical use in real world scenarios, nor publicly accessible without subscription to journals. JEOE tackles this limitation by developing collaboratively online 'open' (licensed under creative commons) documentation and processes. Finally, JEOE considers every ontological activity as a form of boundary setting contributing to define, set and enforce boundaries. Two of the key contributions of JEOE are 1) a strong emphasis on stakeholder analysis and management, and 2) implementation independence, respectively illustrated in the next two paragraphs:

## 1.2 Stakeholders

A diverse stakeholder basis is necessary to a balanced mix of views and sustainability of ontologies, especially their use and long term maintenance. OE consists of a specialized set of activities, requiring depth and breadth of understanding, knowledge and skills in a variety of fields. It is becoming increasingly important to broaden the stakeholder base and to make this process accessible to as many participants as possible, but not at the expense of validity and 'ontological rigour', although even validity and rigor depend on where certain boundaries are set. Furthermore, stakeholders bring onto the development table a much needed socio-technical perspective, of which people and environments are important elements.

## 1.3 Implementation Independence

Ontologies, intended as shared conceptual schemas, should be understood and to some extent even manipulated by a variety of stakeholders with different skill sets (not necessarily skilled in the implementation language of choice) and used as part of sometimes heterogeneous architectures, JEOE advocates clear separation between ontology design and its implementation. Implementation independence is a well established principle in systems science, and it constitutes one of the core features of systems architectures. In the early days of computing it was advocated by Childs [1]. Codd and Dates applied it to the relational model as 'data independence', [2], [3] and Model Driven Architectures translates the principle into 'platform independence' [4]. In Ontology, Tom Gruber [5] advocated implementation independence with the notion of 'minimal encoding bias' an important principle that however is sometimes disregarded by younger generations of ontologists. Finally, JEOE advocates and guides the development of an ontology by ensuring that scope, goal, and most of the conceptual modelling tasks are carried out in advance of any coding, and by not constraining the choice of encoding to a single language: in JEOE it does not matter what formalism the ontology is eventually represented with, it could be OWL, RDF, DL foo, or any encoding of choice. An ontology can exist as a conceptual model independently of its implementation and encoding. This detail is important to counter a trend where ontologies are developed straigh into the ontology editor of choice, and available only as .owl files with little or no other documentation.JEOE provides a compact integrated and 'lean' summary overview that can be used as a learning tool and a streamlined process to guide not only development, but also, and especially, the 'management' of ontology development projects. At some point someone who may know little or nothing about ontologies, will have to budget, cost, staff, and pay for ontology development projects. JEOE provides some background to assist with information necessary to the budgeting scheduling and administration of ontology related projects. Not always however what is currently included in JEOE it going to be by sufficient by itself to solve all ontology development challenges: in some cases more than 'just enough' OE can be required. .

# 2. A SYSTEMIC WORLDVIEW

A system can be defined as 'a collection of components organized to accomplish a specific function or set of functions'. The term *system* encompasses individual applications, systems in the traditional sense, subsystems, systems of systems, product lines, product families, whole enterprises, and other aggregations of interest. A system exists to fulfill one or more missions in its environment. [IEEE 1471]. For a systemist, mostly everything that exists can be viewed as a system, or part of one. JEOEs is developed under a systemic worldview. In Creswell [9] worldview means 'basic set of beliefs that guides action, however in our research a *worldview* is 'preferred conceptual model of the reality under observation', similar to what is referred to a *paradigm*. Cresswell identifies four main worldviews involved with mixed method research .Our worldview can be defined as 'systemic', and to some extent 'transformative', intended respectively as specializations and refinement of Creswell's pragmatic worldview, whereby 'systemic refers to worldview that identifies the widest possible boundary (considers the system as a whole), and transformative as a worldview capable of delivering systemic change. JEOE considers ontology as a 'system' with many dimensions: for example it can be a cognitive tool (as a conceptual mental model), as well as a knowledge representation mechanism in the artificial intelligence sense. A systems engineering approach is useful to provide consistency and coordination for the many disparate ontology development activities when they are considered as a 'system' - but more importantly, the system approach helps to keep in mind that an ontology is part of a greater whole, which is the system ontologies are developed to serve (the web) and their deployment in any target environment and their intended socio-technical usage with all their implications.

## 2.1 Principles of Systems Engineering

There is no single, definitive shortlist of what 'principles of systems engineering' consist of, however some of the SE principles have been instrumental in the development of agile methods [7], namely:

- Start with Your Eye on the Finish Line (be pragmatic about an ontology project)
- Stakeholder Involvement is key

JEOE also matches neatly the requirements identified in other recent ontology related methods such as DILIGENT [6] namely Decentralization, Partial Autonomy ,Iteration, Non-expert builders .

# 3. JUST ENOUGH APPROACHES

The software engineering practice is relatively new. It is interesting to notice how as systems have gradually become more complex, sophisticated and larger, code has become more efficient (more functionality per line of code), and software development methodologies have tended to become leaner, more agile. *'Just enough'* approaches started to emerge when personal computing was just a prospect, and 'structured' approaches to systems analysis and design promised to capture diagrammatic and schematic representation of essential aspects of systems components. This was a response to the growing demand for non formal (mathematical) way of expressing systems requirements and functionalities, and that would be more articulate than pure narrative description of the systems functionalities. That's when structured charts, data flow, data model diagrams, and data dictionaries, started to become in use, with the aim to capture and represent what most counts of a given set of design and modelling activities. Software design is considered an engineering activity, but in many ways it is still a bit of an art. So is ontology development. Directly or indirectly structured approaches have had a huge influence on modern engineering. In a first person account Tom De Marco [10] writes about the events that led him to devise structured analysis, with successes and failures. He admits that "Important parts of the Structured Analysis method were and are useful and practical, as much so today as ever. Other parts are too domain-specific to be generally applicable. And still other parts were simply wrong". Structured approaches are typically viewed as 'top down' or 'waterfall' methods, which perhaps explain why despite their inherent propensity for iteration, can end up not always being considered 'agile'. One of the most successful interpretations of the structured approach to data analysis is to be found in *'Just enough structured analysis',(JESA)* [12] a book that offered 'the best of' structured analysis in simplified form. The JESA wiki states: "Today, we're too busy to spend much time thinking about anything, and we're also far too busy to read more than a couple hundred pages of the bare essentials on any topic. What we want is "just enough" — enough to give us the basic idea, enough to get us started, enough to give us a grounding in the fundamentals." This was true in the early days of data modeling, as it remains true today. 'Just enough' philosophy has left a mark in much current IT literature and practice. Today, principles of 'just enough' thinking and structured methodologies inspire leaner yet robust approaches in many fields. Now a 'just enough' approach is called for in OE, which is a discipline in its own right, (and to some, a cult), and has grown into an immense body of knowledge which is difficult, if not impossible, to absorb and process by the average IT team working on time and budget constraints. In addition to adopting *diagrammatic notation*, a structured approach contributes to consolidate the notion of *abstraction* into ontology development practice. In data modeling, abstraction is what allows to identify and group information assets based on generic common characteristics that exist independently from their time/space representation. Abstraction is adopted in knowledge modeling, as well as in Object Oriented Design, Unified Modelling Language, Integrated Development Framework. Structured methods rely on the notion of *'single abstraction'* mechanism [13], which consists of extracting a top level view of different aspects of the system, forming the basis for functional decomposition', the technique that drills into top level functions, and breaks them down into smaller functions, while preserving the representation of other functional aspects of the systems such as inputs, outputs, controls and other mechanisms. Diagrammatic methods such as UML, for example, are used as form of ontological notation, although it is sometimes argued that such diagrammatic notations may not have the 'expressivity' required to represent all of the essential ontological formalism, such as axioms. Patterns, also known as 'design patterns' are a modeling technique that has started to become adopted in OE. [14]. Techniques such as decomposition – as we know it in functional and/or task decomposition – are sometimes also adopted in some cases to ontology development, as in the DOGMA [15] approach, which decomposes an ontology into an Ontology Base (set of atomic predicates), and a Commitment Layer (Rules). So various techniques for structuring and abstracting knowledge al already adopted. But the learning curve is steep, reality is infinite, and ontology modeling could go on forever. A 'just enough' approach is intended to inspire practitioners to adopt what's needed from wherever they can get it from (even by mixing and matching different methodologies for example) and to set aside the rest. Simplicity and minimalism are golden rules for elegance in any design discipline, although they should not be traded against reliability and robustness, as too rushed oversimplification can also lead to undesirable weaknesses. What may well be 'just enough' for one project, may not be enough at all for another.

## 3.1 Methodologies, an overview

OE consists of methods for the design and implementation of ontologies in the context of IT, which are generally conceptual and semantic models devised to support various intelligent functions, including information and retrieval in network supported and web based environments. Many aspects of OE methodologies are similar to software and system development ones and typically revolve around a 'life cycle'. Over recent decades countless such methodologies have emerged, often evolving organically out of each other. They can be compared by evaluating parameters that they may have in common, for example with Knowledge Engineering methodologies, the detail of their specification (as opposed to being just an outline, as JEOE intends to be), whether it supports a particular knowledge representation formalism - say frame rather than rules, for example- and whether and to what extent they are application-dependent. Some methodologies are built in to ontology editing platforms another historical factor of comparison has been their conformance to software engineering standards, such as the IEEE 1074-1995. [16]

**Commonly Known Methodologies:**

IDEF5
TOVE
DODDLE
CLEPE/AFM
Cyc method
Mike Uschold and Martin King's method
Michael Grüninger and Mark Fox's method
KACTUS
METHONTOLOGY
SENSUS
On-To-Knowledge
Onto Clean
DILIGENT,
HCOME, OTK methodology,
Ontology Development 101
CO4,
KASquare,
DOGMA (AKEM)
SEKT,
OnTo Knowledge(OTK)
OntoClean
BORO
DILIGENT

Information systems become larger and complex by the day, and ontologies have become necessary to support their integration and management. One of the tangible benefits that they provide, is the facilitation of knowledge reuse and communication, which improves the quality of documentation and thus contribute to reduce defect injection, and to the reduction and containment management and maintenance costs of any ontology. The proliferation of methodologies does not resolve the challenge of balancing theoretical competence – such as the in depth specialist knowledge of academics, and the more pragmatic need for optimising efforts and resources, a priority for organizations, managers and to some extent engineers. Who makes the decisions in an OE project, what budget to allocate, what processes and tools to adopt, are critical factors to the success or failure, and somehow novel territory where managerial competences are limited. Additionally, when it comes to knowledge acquisition, usability, and management of ontology users still perceive many weak spots in methodologies, and consider them not yet sufficiently mature and do not adequately meet their requirements. The paper 'OE, A Reality Check'[17] make the point that "OE research should strive for a unified, lightweight and component-based methodological framework, principally targeted at domain experts". JEOE constitutes a step in that direction.

# 4. JEOE ESSENTIALS STEP BY STEP

Traditional OE methodologies, with a few exceptions, tend to be based on a waterfall approach. In JEOE, the sequence of activities is not strictly prescribed, just recommended, but the emphasis is on iteration. Despite the multiplicity and variety of methods that have become available, ontology development is not a 'one size fits all' practice, although sound principles and good practices are generally universally applicable. Any project needs to a certain extent be ad hoc , and sound enough to guarantee best use of resources, reliability and stability of the output. It must also be agile enough to adapt to rapidly evolving circumstances, requirements and digital environments. It is up to practitioners – thanks to a mixture of wisdom and know how - to draw the line and decide how far is far enough. The reminder of this paper introduces the main JEOE steps summarised in the list below:

[1]. Identify stakeholders, outline stakeholder profile
[2]. Define the purpose of the ontology (emphasis on representation/indexing, problem solving/reasoning)
[3]. Outline requirements
[4]. Identify and survey existing knowledge sources and existing ontologies, elicit existing knowledge Assess why the existing knowledge resources do not meet the intended user requirements, update the requirements with the output of activities above [iterative
[5]. Scoping ontology (defining the boundaries and level of granularity, according to goal and stakeholders requirements) Update the requirements [iterative]

[6]. Devise and implement quality assurance plan Add quality parameters to the requirements [iterative]
[7]. Define the field of competence to identify the knowledge boundaries (competence questions) Match the field of competence with the knowledge sources
[8]. Define the ontology artifacts: Vocabulary
   - Identity concepts/entities/classes
   - relations, axioms
   - Refine and map vocabularies to artifacts
[9]. Transfer concepts to ontology language representation: Select knowledge representation formalisms and annotation depending on stakeholder requirements, scope and goal
[10]. Deploy/systems integration (modular, incremental)
[11]. Testing Evaluation quality monitoring competence assessment
[12]. Publishing
[13]. Maintenance/ Reuse

## 4.1 Identifying Stakeholders

Before coding, some level of analysis and design is always advisable. Depending on the domain, target functionality and desired degree of precision, this process, and the set of requirements that results from it, can be tightly or loosely specified and carried out, but it is never completely casual, and requirements should not be plucked from thin air (as it sometimes happens). They must be elicited from stakeholders, using appropriate requirements analysis techniques. The broader category of 'stakeholders' in current systems design is preferred nowadays to the narrower category of 'users'. Stakeholders include users, but also sponsors, investors, technology providers, industry associations, standardization bodies, and other people and roles not necessarily identified at planning stage. A stakeholder is anyone actively involved in the ontology development and its intended use, and anyone whose interests may be affected by the development of such an ontology. Likely to end up being a large and very diverse crowd, which is what should make the process of OE fun, but that unfortunately can cause struggle and waste of resources. This is because some stakeholders come from traditional environments where their role and point of view is never challenged. But the narrower and more 'authoritative' the stakeholder base, the narrower its scope, and to some extent its ability to gain 'acceptance', therefore its propensity to reuse. The 'stakeholder structure' or 'base' should be identified and profiled at early stage in any ontology development project, it should be kept well involved throughout the subsequent phases, to avoid 'disenfranchisement' which can result in lower level of collaboration among different project contributors. A stakeholder analysis process consists of identifying stakeholders, such as persons and roles, and the organizations they belong to, and cluster them according to shared parameters (common goals, interests, requirements, tasks responsibilities.

'Stakeholder management' should be carried out to leverage the patterns and dynamics of stakeholder groups, their goals, motives and commitments, and to create and sustain the collaborative momentum that can fuel an ontology development project itself. One of criticisms that semantic web technologies have faced in recent years, is that they were not really designed with users, or usefulness, in mind. A lot of (publicly funded) time and money has been spent to develop tools, platforms and environments that were experimental, and satisfied a particular curiosity of a researcher or to support a theoretical point of view. Nowadays, especially in tightening economic conditions, there is increasing demand for justification, and for adoption of good practices.

## 4.2 Purpose/Goal

A goal is intended as a tactical, precise, measurable target achievement, while a purposes is overall scopes tend to be strategic. Ontologies can be used for a variety of goals and purposes, and once an ontology is in place, and developed following to good practice, it can be used and manipulated without restrictions, even for a purpose different from the one that it was intended initially. Ontologies however are constantly undergoing refinement, and if they are not, they should, at least to some extent. Having a clear goal for the usage and application of the ontology, will help to guide its development, and concentrate the efforts to fulfill the priority requirements. Ontologies are specifications of a consistent and explicit view of reality, but both the view, and the reality they represent, change, and this change must be tracked throughout development. Examples of goals are scattered everywhere in OE literature and include:

- Support a process execution within a system
- Improve the efficiency of reasoning
- Consolidate and harmonize existing data/information
- Provide an abstract, more schematized view
- Create a consensual, unified view that can serve as synthesis of different views
- Provide a formal specification
- Support integration of data, applications, and systems to help minimize design and planning errors caused by lack of domain knowledge

Clearly establishing the purpose helps to understand what kind of ontology needs to be developed - for example a content ontology for reusing knowledge, a communication ontology for sharing knowledge, an Indexing ontology for case retrieval and Meta-ontologies for increased knowledge representation. But a 'good' ontology is going to be useable for almost any purpose. Examples of functions that can be supported or even fully automated, using an ontology [18]:

- Consistency checking ( properties and value restrictions)
- Auto Completion of information partially provided by users
- Interoperability support (shared conceptualization)
- Support validation and verification testing of data (and schemas)
- Configuration support — class terms may be defined so that they contain descriptions of what kinds of parts may be in a system.

A simple example of goals can be: "the ontology serves as a means to structure and verify the validity of any information set (from a restaurant menu to the diagnose of a complex medical symptom), or "*the ontology serves as a set of parameters for integrity constrain within a given process*". The ontology 'goal' ideally emerges from agreement/consensus of the stakeholder basis, the members of which probably spend a lot of time arguing, among other things, about what goal should take priority over another, or how two goals may be conflicting. These are generally necessary labor pains for any OE project. What sometimes happens, is that when different stakeholders cannot agree over the priority of a goal for an ontology, this naturally serves as the ontology 'split point', where subsets of stakeholders should dedicate themselves to develop one particular aspect of the ontology, according to their priority and preferred goal. (This is also true for any other decision where consensus cannot be achieved). The entire stakeholder base should then only agree on common parameters, for the purpose of facilitating the merging, interoperability and integration of the respective outcomes at a later stage. Reconciling different stakeholder views can be managed using standard brainstorming and knowledge sharing techniques, supported by mind mapping tools, or by any of the tools designed for this purpose, many of which are free or open source. Framing such goals into set specifications could be useful , but should be done so with a degree of flexibility. Goals should also be periodically revised, as the project requirement and context may change.

## 4.3 Requirements

An ontology may well serve more than one purpose/goal but it will always help to structure, analyse, communicate and share and reuse their knowledge about a particular domain, task or process. An ontology is not strictly speaking 'software', however it is generally represented, used, queried and manipulated using software artifacts. Much of requirements engineering practice applied to software development can be adopted in OE. For example it is also possible to distinguish to some extent between *system/functional* requirements, when they solve a particular problem, provide a functionality, enforce a constraint, and *user* requirements, when they are designed around user tasks and need, although often the two often overlap. Some ontologies may originate from the need of a stakeholder, or a group of stakeholders, to define and specify a given notion, concept, domain, problem, field of action that they are working on. It can be argued that not every ontology needs a set of requirements, and that sometimes ontologies just 'happen' as the result of pulling together the cognitive artifacts of a given task or profession or team. But considering that scientific and technical domains are complex, and information is becoming more challenging to manage, and that the validity of information technology artifacts depends on their accuracy, when developing ontologies, organizations do so by allocating resources. Be it employee time, skills or equipment, specifically to address given problems, the returns on investment are carefully weighted against results. *Did the ontology solve the problem it was developed for? To what extent? At what cost?* Whether an ontology meets its requirements will be an important parameter to measure the success or the failure of a project, and to gauge quality evaluation. . Quality targets should be included in the ontology requirements. It is important that the ontology complies with the expected quality parameters, as it is important that it answers the 'competency questions'

### 4.3.1 Requirements Input

The requirements for an ontology should be developed by taking into account first and foremost the stakeholder input, possibly compiled following some structure in relation to the goal and purpose of the ontology, as well as to the other stakeholders inputs. Additional ontology requirements should be derived from an analysis of scope, granularity, quality standards, implementation languages and environments, discussed next, and which must be decided throughout development, and often in parallel with other activities, and integrated at each iteration point. Some requirements are likely to be 'fixed', that is, they cannot changed, and others will emerge and evolve during development, in which case both 'sequential' and 'iterative' approaches to compiling the requirements for the ontology can be combined Among desirable top level requirements already discussed elsewhere, some generics to keep in mind are:

- declare explicitly what high level knowledge (upper level ontology) it references,
- declare explicitly what kind of reasoning/inference supports and it is based on.
- be accessible to all the agents/agencies (this means shared, viewable, understandable)
- be 'acceptable' to all the agents/agencies from the different perspectives, in terms of point of

view, culture, language, conformance to policy and protocols

- 'usable' in terms of compatibility with local information systems used by agents/agencies

## 4.4 KNOWLEDGE SOURCES

Many ontologies are published on the internet, although not all of them are publicly available and accessible, and sometimes they are protected by intellectual property rights. Whatever ontology is needed, chances are that one already exists, or is being developed, by someone else, however, other existing ontologies may not necessarily comply with the set of requirements of the given project at hand. Knowledge drives decision making and behaviours (can be seen as a form of organizational energy the transfer and exchange of which leads to transformation) [27] however not all knowledge is 'factual', and only when facts are supported by, and ideally linked to evidence (provenance) that they can be disambiguated from beliefs and opinions [25] (Figure 1)

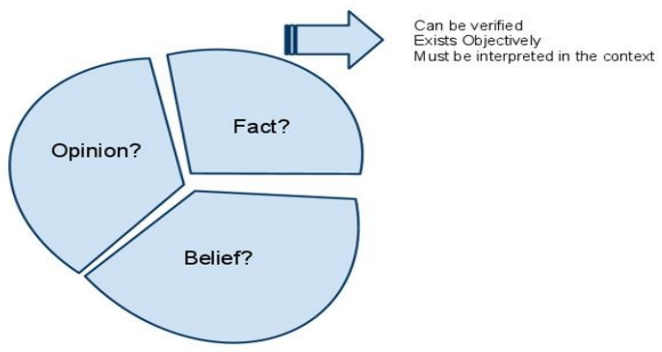

Figure 1. Fact, Opinion or Belief? (Di Maio, 2010)

One of the good principles of knowledge reuse prescribes to source and recycle what already exist, if possible, to the extent that it is possible. The obvious 'cost of reuse' rule also applies, where if the cost of reuse is higher (in terms of acquiring a license, for example, or in terms of decoding an ontology that has been heavily committed to a formalization), then the choice not to reuse is fully justified. Although it may not always be possible or convenient to reuse existing ontologies, it is good practice to acknowledge and reference them for completeness. The knowledge that constitutes the foundation of an ontology, is always grounded, elaborated and derived from knowledge that was existing before, which in some cases is remixed and reinterpreted to suit a novel requirement. Other forms of structured knowledge repositories and that are not ontologies are encyclopedias, libraries, indices, dictionaries, archives, and scientific and technical publications tend to reference and include references to the body of knowledge (BOK) of any given domain. A lot of knowledge is scattered in various non structured forms, An ontology aims to map, synthesize and resolve the conflicts that exist within the knowledge sources that constitute the body of knowledge of any given domain. When developing an ontology, all of the above knowledge sources should be considered, including other existing ontologies .When the analysis and summary of existing knowledge sources does still not satisfy the requirement, and does still not provide the answer to the questions being sought, or does not support the desired functionality (for example being coded into a particular language) then the scope of the work to be done becomes clearer and better documented. The result of this evaluation would reinforce or modify the initial requirements and specification, taking into account what can be reused. Given the messy state of affairs of information and knowledge sources today (much of which is outdated, poorly accessible etc) , then writing up things from scratch can sometimes be quicker and more efficient, and more likely to conform to the needs in hand. But existing knowledge should always be audited and inventoried in an OE project. How much one should try to reuse what was there before, and how much one should reinvent a new, is often a 'just enough' type of decision.

### *4.4.1 Knowledge Sharing and Reuse*

At least two research directions have motivated (and funded) OE research developments of recent years: one is the need to provide knowledge representation mechanisms that are 'shared', that is commonly accessible and understood, so that knowledge can be reused, and knowledge flows optimised. Another is provide mechanisms for artificial intelligent agents to perform reasoning functions. The latter conflicts with the former when intellectual property rights such as patents are also the intended, albeit secondary purpose, for research. It is worth remembering that ontologies in general are devised to facilitate the sharing of knowledge, whether among restricted or open agents, whether these are human or artificial. Generally, for knowledge to be reused, it needs to be shared. Ontologies provide sets of parameters for knowledge sharing, and rely on the assumption that all the constructs and artefacts are shared. Across large domains, taking into account the diversity of disciplines, paradigms, axioms, vocabularies, uses, standards, practices, and despite many years of OE research and practice, knowledge sharing good practices (such as shared vocabularies) are still not adopted, or only marginally so, outside the relatively small knowledge engineering community. A knowledge audit can provide an overview of the (explicit) knowledge, and its qualitative and quantitative characteristics, helping to identify the location where it resides, as well as other information such as people and roles involved in their creation and maintenance, and other organisational processes associated with it. In related research a Knowledge Audit Framework [19] is devised for the purpose of facilitating systematic auditing as well as sharing and reuse, of artefacts and schemas.

## 4.5 SCOPING THE ONTOLOGY

Reality is complex, resulting from of the dynamic combination of what exists, which is not always observable, and its underlying dynamics which are only partially 'knowable'. Causes are often imponderable. Ontologists describe these layers as 'levels of reality'. A simple yet effecive example is provided by the decomposition of the levels of reality of a simple nut and bolt [20], whereby a compound object is made up of components which in turn are made up of elements which in turn are made up of particles. At what level of reality is the target ontology going to be pitched? That needs to be specified and defined very early, and the outcome of this decision is also going to constitute part of the specification document. OE addresses the different levels of reality by specifying ontologies which define reality at the appropriate level of granularity, depending on what they are describing, Such distinction is reflected in the differentiation between UPPER, DOMAIN, APPLICATION, TASK, or PROCESS ontology. Segmenting the ontologies according to their scope simplifies tremendously the effort of referencing existing knowledge, as different axioms and paradigms may rule the different portion of reality being addressed, and to specify their intersection, for example, where physics (the science that studies elements) interacts with ergonomics (the science that studies how people work), making the ontology that is being built 'grounded', therefore more stable and reliable and reusable in the future. The ontology specification document should include the degree of formality, which addresses, which is determined either by the existence and weight of axioms, or by the degree of formality of the ontology representation language (OWL is said best to support axiomatic formalism, etc). But one should remember that even the most informal ontological statements rely implicitly or explicitly on the existence of at least one, axiom.

## 4.6 QUALITY ASSURANCE

One of the established methods to evaluate quality of artifacts, is to develop a 'quality model', which should be done during the early stages of the ontology development, and serve as guidance throughout the project. Quality Models are developed upfront, and used as target parameters throughout the development, evaluation and testing, Just to reiterate, while testing is generally done at the end of the development (or of each iteration) , quality evaluation can only be performed if quality parameters are set upfront: quality models contain patterns of qualitative and quantitative measurements of various aspects.

The quality of an ontology is sometimes measured across two dimensions: its accuracy and its comprehensiveness,[22] corresponding to the notions of precision and recall in search technology. Almost the entire range of standard testing techniques used in programming consistency integrity, validation, redundancy can be applied to test the validity of an ontology. A good summary of quality evaluation criteria for ontology can be found in [21]

## 4.7 COMPETENCE

In addition to using known software, project quality, and evaluation techniques, ontologies rely on special, specific tests: competency tests checks (also known as "competence test"). The competence domain of an ontology indicates the knowledge field that the ontology represents (or should represent). In order to answer any given competency question, the ontology should contain all the knowledge parameters necessary to formulate a correct answer for that question. Competency checks are sets of questions used to determine the competency of an ontology. It is useful to develop these competency questions from stakeholders input and throughout the project lifecycle. Quality planning is done up front, but the quality model should be updated dynamically and iteratively throughout the project. Different tests can be set up to verify the validity and quality of each part of the ontology, each process within the ontology development, and carried out correspondingly at each step.

## 4.8 DEVELOPING THE ARTIFACTS

An ontology is defined by the boundaries that constitute it. This boundary setting starts early in development, as stakeholder and goals are in themselves the first set of boundaries. However the real definition of an ontology tightens up when getting down to devising its artifacts. Much has been said about what constitutes an ontology. Conceptualisations, models, schemas, representations, frameworks. An ontology may take a variety of forms, but it will necessarily include a vocabulary of terms and some specification of their meaning, such as definitions and an indication of how concepts are interrelated, which collectively impose a structure on the domain and constrain the possible interpretations of terms [Uschold et al.].

***VOCABULARIES:*** Encyclopedias dictionaries, thesauri and vocabularies are fundamentally lists words and their definitions, which can include grammatical, phonetic and etymological annotations. Thesauri are vocabularies where the semantic association between terms are mapped, while glossaries are alphabetized lists of terms with definitions usually appended at the end of documents or reports. Information systems adopt vocabularies to support design and documentation, 'data dictionaries' for examples are used to list the entries used in a database. Vocabularies are at the core of ontologies, to the point that sometimes they are referred to as being the ontology itself. They list terms that declare and represent every concept, relations, function and axiom, the more an ontology is formal, the stricter the definition of its vocabulary terms. In an ontology, the vocabulary has more than one function, serves an index, and a directory of content. Generic vocabularies can contain more than one definition for each term, but controlled vocabularies do not, as they allow only one definition per term and explicitly enumerated (numbered) terms, which must be unambiguous, and non redundant. Vocabulary creation is both an art and a science, which leverages principles of information and library science, the core notion however is to keep track of the words (lexons) used in the ontology itself, as well as in the discussions that lead to the development of an ontology. How to combine different kinds of vocabularies to make the most of organizational knowledge is currently being researched. In addition to vocabularies, which are used to 'name' the artifacts, components are necessary for an ontology to take place: concepts, relations, axioms, discussed briefly below

**CONCEPTS:** Concepts are fundamental to our ability to think, express, represent and communicate, however, defining unambiguously and with certainty what constitutes a concept, is rather tricky, and pushes IT practitioners toward the realm of philosophy, where they can get easily lost and will never come back to IT proper. But that's a challenge of OE. Concepts can correspond to things, but also to 'fuzzy clouds' of ideas and notions identified by words and related to a certain thing or subject. And even when referring to tangible things, concepts can be abstract, and difficult to be captured. The nearest technique that can be compared to conceptual modeling, is entity modeling, or class modelling. Concepts can be broadly divided into cognitive artifacts that support categorization and communication, and are necessary to support human and artificial thinking and reasoning. The purpose of ontologies is to make them explicit and represent them so that they serve a variety of purposes, namely the intended goals. Conceptual categories and thoughts are closely related to language. A concept model can be used to complement and extend a functional data model, In OE, concepts can be modeled following 'formal concept analysis (FCA)',

**RELATIONS:** Most views of reality are perceived as dynamic combinations of things and entities, kept together by correlations and dependencies. In models of reality, such as ontologies, the semantic interdependence between a thing and another is considered a relation. Relations are the cognitive counterparts of dependencies in the real world. There are different kinds of relations, and there is no single theory that studies them all. A fundamental representation of a relationship between two concepts is a mathematical structure denoting it as a set mapping between the instances belonging to the two concepts. These mappings might be characterized along the dimensions outlined below

- Arity: Typically binary relationships are of most interest, but relationships can be of arbitrary arity, i.e., we could have 3 or more concepts participating in a relationship.
- Cardinality: These constraints are characterized in one of the following ways: 1-1, many-1, 1-many, or many-many. A more generalized way of representing these cardinality constraints is using a pair of numbers that specify the minimum and maximum number of times an instance of a concept can participate in a relationship. This is a very useful technique for n-ary relationships and also captures partial participation of concepts in relationships. 1-1 and many-1 relationships are functions which can be exploited in various ways.
- Direct v/s Transitive Relationships: Some entities might be directly related to each other via their participation in a common relationship, or might be related transitively to each other via a chain of relationships.
- Crisp vs. Fuzzy: Most of the current modeling approaches view relationships as crisp, i.e., for an n-ary relationship, instances of n concepts are either part of a relationship or not (e.g., is-a, part-of relations). In the case of fuzzy knowledge [Zadeh65], the extension of a relationship may be viewed as a joint probability distribution on the concepts participating in a relationship. For example semantic similarity (i.e., proximity) between two entities is an example for fuzzy relations.
- Properties vs. Relations: Properties are special relationships where the ranges of a relationship are values of a data type (e.g., dates, age) as opposed to instances of a concept.
- Structural Composition: Relationships can either be composed (if they are functional in nature) or combined using join operations to create new relationships and associations based on existing relationships.

Computational techniques can be used to identify, discover, validate and evaluation relationships within any given knowledge and reality schema 'Relation' is a canonic class of any ontology. It is characterized by substantial properties and formal attributes. Of the material properties, there are their reality, nature and type and direction of dependency. Of the second, there are transitivity, symmetry, reflexivity, and n-ary, or cardinality, terms, or tuples, of domains, elements, components, or arguments). [25] Three things are of importance

1. the components of relations are of the same kind and sorts, objects, persons, qualities, quantities, times;
2. ordering of relations, their direction, a triadic 'giving', tetradic 'paying' or triadic 'betweenness';
3. the key sense of relationship is represented by the graph, indicating its nature and kind: if it's causal relation,

temporal relation, spatial relation, semantic relation, logical relation, etc.

When represented in the context of linguistic representation, relations are called 'lexical relations' that in turn can be of many kinds. Generally known are 'taxonomic relations (Synonymy, homonomy) or non taxonomic [22] An sample set of fundamental ontological relations can be viewed in the Relation Ontology, from the OBO Foundry Project [23].

**AXIOMS:** In ontology, axioms serve to model sentences that are 'always true', and they are used to verify the consistency of the ontology, as well as the consistency of the knowledge stored in a knowledge base. Axioms are required in support of any logical statement, and the main distinction that should be remembered is that In traditional logic, they are considered self evident and true, while in mathematics, logical axioms are usually statements that are taken to be universally true. Outside logic and mathematics, the term "axiom" is used loosely for any established principle of some field Axioms translate into constraints, which in turn can be considered as the logical boundaries of an ontology. They can be transformed and mapped easily directly into rules. If an axiom maps to the rule, then consisting parts of an axiom map to the consisting parts of a rule. [24]

The mapping follows:
- ontology axiom → rule
- axiom statement → rule clause
- statement concept → entity in a rule clause
- statement relationship → relationship in a rule clause

## 4.9 IMPLEMENTATION

The implementation stage starts with transferring what has been designed on paper using graphical notation, (say using bubbles and arrows) to ontology language representation: It consists of selecting knowledge representation formalisms and annotation depending on defined stakeholder requirements, scope and goal . According to computer scientist Tom Gruber: When we choose how to represent something in an ontology, we are making design decisions. To guide and evaluate our designs, we need objective criteria that are founded on the purpose of the resulting artifact, rather than based on a priori notions of naturalness or truth. The main principles of OE, as devised around 30 years ago by Gruber [5], still largely stand. They are summarised below:

1. **Clarity**. should effectively communicate the intended meaning of defined terms. All definitions should be documented with natural language. [...]
2. **Coherence**. should be coherent: that is, it should sanction inferences that are consistent with the definitions. [...]
3. **Extendability.** should be designed to anticipate the uses of the shared vocabulary. [...]
4. **Minimal encoding bias**. The conceptualization should be specified at the knowledge level without depending on a particular symbol-level encoding. An encoding bias results when representation choices are made purely for the convenience of notation or implementation. Encoding bias should be minimized, because knowledge-sharing agents may be implemented in different representation systems and styles of representation.
5. **Minimal ontological commitment**. should require the minimal ontological commitment sufficient to support the intended knowledge-sharing activities [...]

Principle 4, *minimal encoding bias*, states that an ontology should be independent of its implementation, whereby the coding should not run the development process. This is something that systems designer and information architects know very well. Concepts can be modeled using mind maps, lattices, Petri nets, and other abstract, diagrammatic and graphical representations. After the conceptual modeling is done, the goals are set, the requirements specified, and the artifacts outlined, it is time to start thinking about actually encoding. Encoding an ontology means to start assigning roles, properties, and values to each term pointing to an artifact. That's when the appropriate ontology representation language is selected unless a specific code set/ implementation option is part of the initial requirement, in which case some encoding decision may well precede other aspects of the development and inevitably influence it. The knowledge contained in ontologies can be expressed with different knowledge representation formalisms and languages capable of supporting logical assertions — from frames to semantic networks and axioms, where the most common formal notation in use is description logic. On the Web, the current standard "grammar" for ontology representation is the Resource Description Framework (RDF) and OWL.To have the ontology exist in a variety of notations and formats, not only in RDF and OWL, means that ontology can effectively be worked, manipulated, and viewed in different environments and not just by ontology editors and Semantic Web browsers of one kind. However, to test and assess the reasoning that an ontology is capable of supporting, having an implementation or working prototype is necessary. An ontology can satisfy competence tests based on paper models; however, it is only when implemented that a system will be able to check to see if it works and if it helps reasoning or totally warps it. Documenting the ontology development process as well as the implementation/encoding process is important. Especially because during testing and evaluation, it will be necessary to be able to identify if errors are actual conceptual flaws in the model or are caused by flaws and errors in the actual implementation and encoding of a correct conceptual model.

## 4.10 Deployment

Having an ontology all done, implemented, and working is probably never going to happen as a single event because it is likely to remain a continuous, staged process. Reaching the end of development process is only a beginning and the completion of an iteration. Using the ontology to serve its purpose means integrating it with the rest of the information systems environment it needs to work with and that can never be defined entirely. It is likely to evolve together with the configuration of the system and infrastructure. An ontology or an ontology module can be released as a compact, standalone, all-in-one artifact that can be zipped up and downloaded as a single file; however, defining what part of an information system or infrastructure that the ontology should interface with and relate to — as well as what level of integration the ontology should support — is another aspect of ontology development that should be planned early and possibly form part of the requirements.

## 4.11 Testing and Validation

In OE, testing should be considered a subset of the overall quality assurance activity plan. The validity of different parts of an ontology must be differentiated, as each and every aspect of the conceptual and semantic model is not necessarily directly related to the quality and validity of the ontology code and implementation. The robustness of the conceptual model and the validity of the semantic artifacts should be verified separately from the code, although integrated tests should also be carried out over both. Almost the entire range of standard testing techniques used in programming such as consistency, integrity, validation, redundancy testing, as well as usability testing can be adapted to test an ontology. The key is to isolate the component/aspects that need to be verified and to set up the corresponding set of tests for each artifact or set of artifacts (sometimes referred to as “units”), as well as carrying out overall performance checks.. An ontology should be tested for accuracy — the ability to support correct inference — in relation to the expected competence range. It should be tested to verify its ability to identify and represent correctly linguistic intensionality and extensionality, for example, where the first refers to the “aboutness” of an expression, and the latter refers to the ability to capture and represent the context for the intended meaning.

## 4.12 Publishing

An ontology can represent the most abstract form knowledge representation of a task or domain. It contains, in distilled form, a big portion of intelligence of whatever is defined. By definition, publishing means making it accessible to others where it can be found, referenced, and used. The first issue is about IP control and what level of public access should be assigned to the ontology. This is not a technical issue, and the level of disclosure of the ontology should conform to the general IP policy of the organization that is developing the ontology. Provided the ontology is for public disclosure and appropriate license is attached to it, publishing generally entails making it available on the Web or other universally accessible repository and making it discoverable. There are certain ontology schemas repositories available on the Web such as Schema Web, but in a linked data model it does not matter where the ontology files are published because a semantic connection between this file and other corresponding objects on the Web would enable and facilitate its discovery

## 4.13 Maintenance/Reuse

In our world of rapid changes, the lifecycle of an ontology is directly related to its ability to maintain its currency. There is only one way that this can realistically be done and that is to support dynamic updates. This may well become possible in the future, but at the moment it’s just a good plan. Ontologies still need maintainors and curators to make sure that they do not become obsolete too quickly. Certain facts are not likely to change very often; however, other things change regularly and need updating. Think of policy, legislation, and security. An ontology will need to be maintained and updated periodically, and this must be set up as an ongoing task that will extend well beyond development. In JEOE we bunch ontology alignment, merging, and reuse under “maintenance” because they all represent an early step in the lifecycle of the next ontology, thus completing a loop.

# 5. OTHER JEOE RESOURCES

Aside from an agile process to guide development in way to maximise stakeholder involvement and reduce the risks of failure by identifying common stumbling blocks and suggesting workarounds and management decisions to avoid wasting resources and getting stuck in dead ends, JEOE aims to point to essential knowledge resources that constitute a minimal core body of knowledge for the practice, offering a minimal and hopefully lucid Even before being consolidated as a set of core steps, JEOE started of an approach, a systems engineering survival guide to ontology development, reflecting an engineers attitude to life and the universe: a determination to get things to work , one way, or another, and characterised by extreme flexibility and an ability to make the right decision at the right time. Engineers, and 'systems engineers' in particular, view everything as a 'system' and tend to tackle every problems or challenge 'systemically'. Its hard to prescribe what that is, as a great deal of heuristics goes into a systemic approach, generally guided by experience. The first version of JEOE was published y Cutter Consortium, whose founders are also the early pioneers of Just Enough principles. A current open online version of JEOE exists as a website designed for future reference and open collaborative development of the same http://www.justenoughontology.co.uk/

## 6. ACKNOWLEDGMENTS

Thanks to those who generously share expertise and resources on the open web, especially Ed Yourdon and Tom De Marco, for inspiring and guiding just enough thinking towards JEOE, Cutter Consortium for publishing the first JEOE version. Knowledge Reuse and Knowledge Auditing Framework related research partly funded by EPSRC Grant Number ,EP/D505461/1 for which the author thanks the University of Strathclyde.